%% file: arxiv.tex
\documentclass[10pt,twocolumn,letterpaper]{article}

\usepackage[pagenumbers]{cvpr} 


\definecolor{cvprblue}{rgb}{0.21,0.49,0.74}
\usepackage[pagebackref,breaklinks,colorlinks,allcolors=cvprblue]{hyperref}

\def\paperID{*****} 
\def\confName{CVPR}
\def\confYear{2025}

\title{An Individual Identity-Driven Framework for Animal Re-Identification}

\author{Yihao Wu, Di Zhao\thanks{Corresponding author.}, Jingfeng Zhang, Yun Sing Koh\\
School of Computer Science, University of Auckland\\
Auckland, New Zealand\\
{\tt\small \{ywu840,dzha866\}@aucklanduni.ac.nz, \{jingfeng.zhang,y.koh\}@auckland.ac.nz}
}

\begin{document}
\maketitle

\begin{abstract}

Reliable re-identification of individuals within large wildlife populations is crucial for biological studies, ecological research, and wildlife conservation. 
Classic computer vision techniques offer a promising direction for Animal Re-identification (Animal ReID), but their backbones' close-set nature limits their applicability and generalizability. 
Despite the demonstrated effectiveness of vision-language models like CLIP in re-identifying persons and vehicles, their application to Animal ReID remains limited due to unique challenges, such as the various visual representations of animals, including variations in poses and forms. 
To address these limitations, we leverage CLIP's cross-modal capabilities to introduce a two-stage framework, the \textbf{Indiv}idual \textbf{A}nimal \textbf{ID}entity-Driven (IndivAID) framework, specifically designed for Animal ReID. 
In the first stage, IndivAID trains a text description generator by extracting individual semantic information from each image, generating both image-specific and individual-specific textual descriptions that fully capture the diverse visual concepts of each individual across animal images. 
In the second stage, IndivAID refines its learning of visual concepts by dynamically incorporating individual-specific textual descriptions with an integrated attention module to further highlight discriminative features of individuals for Animal ReID. 
Evaluation against state-of-the-art methods across eight benchmark datasets and a real-world Stoat dataset demonstrates IndivAID's effectiveness and applicability. 
Code is available at \url{https://github.com/ywu840/IndivAID}.

\end{abstract}

\section{Introduction}
\label{sec:intro}

Biologists are usually interested in concentrating on a large group of individuals due to their unique roles or behaviors, making it essential to investigate and collect data from the population~\cite{korschens2018towards}. The capability to reliably re-identify these individuals is crucial for extending our understanding of ecosystem functionality, community dynamics, population ecology, and behavioral studies~\cite{schneider2019past}. Furthermore, precise tracking of animals in the wild allows ecologists to approximate the population of various species and monitor the health status of endangered species, demonstrating wildlife conservation. However, identifying these individuals can be challenging due to the subtle visual differences among individuals and the unconstrained environment~\cite{korschens2018towards, ravoor2020deep}. Therefore, constructing a reliable and autonomous Animal Re-identification (Animal ReID) framework is necessary for both humans and animals.

Traditional Animal ReID methods used by ecologists, including tagging, scarring, branding, and DNA analysis from biological samples, are effective but have notable weaknesses. These methods are labor-intensive and time-consuming, sensitive to sensor failure, challenging to scale for large populations, and limited in their ability to assess animal-environment interactions~\cite{schneider2019past}. With the advancement of computer vision and deep learning, incorporating computer vision into the Animal ReID framework has shown a promising direction~\cite{ravoor2020deep}. Some state-of-the-art Animal ReID methods have been proposed in deep learning Animal ReID.

Within the computer vision domain, Convolutional Neural Networks~\cite{lecun1998gradient} stands out as the prevalent architecture for ReID tasks, effectively mapping images to feature representation within an embedding space to minimize intra-class distances while maximizing inter-class separations. While CNN-based methods for discriminative feature representation learning in Person ReID tasks have demonstrated significant success on benchmark datasets, their practical application remains limited~\cite{li2023clip}. Specifically, CNN-based methods usually focus on minor local image features that are potentially irrelevant image regions, leading to a lack of robustness and generalizability.

In contrast, the Vision Transformer (ViT)~\cite{dosovitskiy2021an} has gained superiority across various tasks, including Person ReID, where it often outperforms CNN by capturing long-range dependencies within images. However, ViTs come with vast parameter sets, requiring extensive training datasets and posing challenges during optimization. Given the relatively small size of most Animal ReID datasets, the full potential for ViT models in Animal ReID tasks is unexploited.

Most existing CNN and ViT approaches utilize the backbones pre-trained on the ImageNet~\cite{krizhevsky2012imagenet} dataset, which limits their suitability for ReID tasks. This limitation arises from the close-set nature of ImageNet pre-training, where the label space remains consistent across training and test datasets. However, ReID presents an open-set challenge, with disjoint label spaces between training and test datasets, introducing a domain shift. 

By leveraging the Vision-Language Models, the invention of vision-language pre-training frameworks like CLIP (Contrastive Language-Image Pre-training)~\cite{radford2021learning}, trained on vastly larger datasets (400 million images versus ImageNet's 1 million), aligns more closely with ReID requirements. CLIP aims to learn visual concepts from the perspective of natural language. Unlike image classification tasks, CLIP associates visual features with corresponding language descriptions. Such a description also refers to a prompt in the Natural Language Processing (NLP) area. The classification process involves comparing image features generated by the image encoder, which takes the original image as input, and text features generated by the text encoder, which takes the prompt as input. Consequently, CLIP's image encoder can capture diverse high-level semantics from textual descriptions, which can be adapted to various tasks.

The main challenge to adapting CLIP to ReID tasks is that CLIP requires a textual description with a specific text label to form a valid prompt. However, in ReID tasks, labels are numeric indices lacking descriptive text labels for images.
Additionally, compared to Person ReID, it is more challenging to develop a generalizable and robust Animal ReID model due to the variations of animals in appearances, poses, and forms in Animal ReID tasks~\cite{schneider2019past, ravoor2020deep}.

To overcome these challenges, we propose a two-stage framework, the \textbf{Indiv}idual \textbf{A}nimal \textbf{ID}entity-Driven (IndivAID) framework, for Animal ReID. In the first stage, IndivAID trains a text description generator to produce an image-specific and individual-specific textual description for each image input while freezing CLIP's image and text encoders. In the second stage, IndivAID fine-tunes CLIP's image encoder by incorporating the individual-specific textual descriptions integrated with an attention module but keeps the text description generator and CLIP's text encoder fixed.

Our contributions are summarized as follows:
\begin{itemize}
    
    \item We present IndivAID, a two-stage framework that utilizes CLIP's cross-modal capabilities to generate textual descriptions for each individual. By adopting the prompt learning strategy, IndivAID trains a text description generator to produce an image-specific and individual-specific textual description for each image. The text description generator is designed to address the numeric index-based labels in ReID tasks. Moreover, each individual's learned textual descriptions consider each image input, which helps overcome the challenge of variations in animal images.
    
    \item All the constructed textual descriptions from the Stage One training are merged into individual-specific textual descriptions with an attention module, enhancing IndivAID's performance in ReID by providing a flexible merging strategy.
    
    \item Evaluation against state-of-the-art methods across eight Animal ReID benchmark datasets and a real-world Stoat dataset demonstrates IndivAID's effectiveness and applicability in re-identifying animals with diverse characteristics.
    
\end{itemize}

\section{Related Work}
\label{sec:related_work}

\noindent \textbf{Deep Learning for Animal ReID.} Konovalov et al.~\cite{konovalov2018individual} utilized CNN with data augmentation technique developed to train an Animal ReID model for re-identifying dwarf minke whales. They designed an Animal ReID system to track the whales's population and investigate the impacts on tourism development. Phyo et al.~\cite{phyo2018hybrid} concentrated on the localized part of cows to extract the black and white patterns on cow's body for constructing a CNN-based Animal ReID model. Similarly, the researchers are dedicated to tracking the health conditions of cow individuals in the farms. Konovalov et al.~\cite{konovalov2018individual} and Phyo et al.~\cite{phyo2018hybrid} proposed closed-set Animal ReID solutions, assuming each individual in the Query Set also appears in the Gallery Set, and all the animal species in the Gallery Set also appear in the Training Set. Nonetheless, Bouma et al.~\cite{bouma2018individual} incorporated ResNet50~\cite{he2016deep} into the dolphin ReID network from an open-set recognition perspective.

\begin{figure*}[!ht]
  \centering
  \begin{subfigure}{\columnwidth}
    \includegraphics[height=0.47\columnwidth, width=\columnwidth]{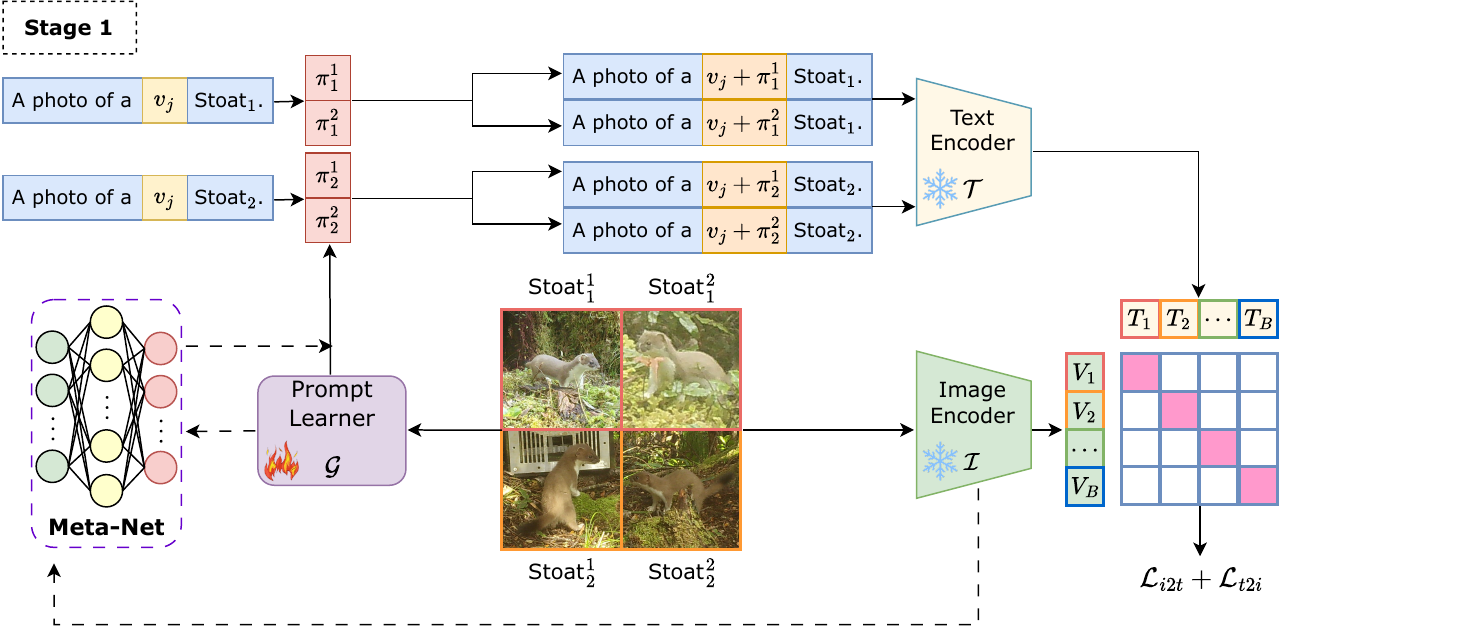}
    \caption{\textbf{Stage 1:} Training of the Text Description Generator, $\mathcal{G(\cdot)}$, to produce an image-specific and individual-specific textual description for each image with a fixed text encoder, $\mathcal{T(\cdot)}$, and image encoder, $\mathcal{I(\cdot)}$.}\phantom{XXXXXXXXXX}
    \label{fig:IndivAID_stage_1_framework}
  \end{subfigure}
  \hfill
  \begin{subfigure}{\columnwidth}
    \includegraphics[height=0.47\columnwidth, width=\columnwidth]{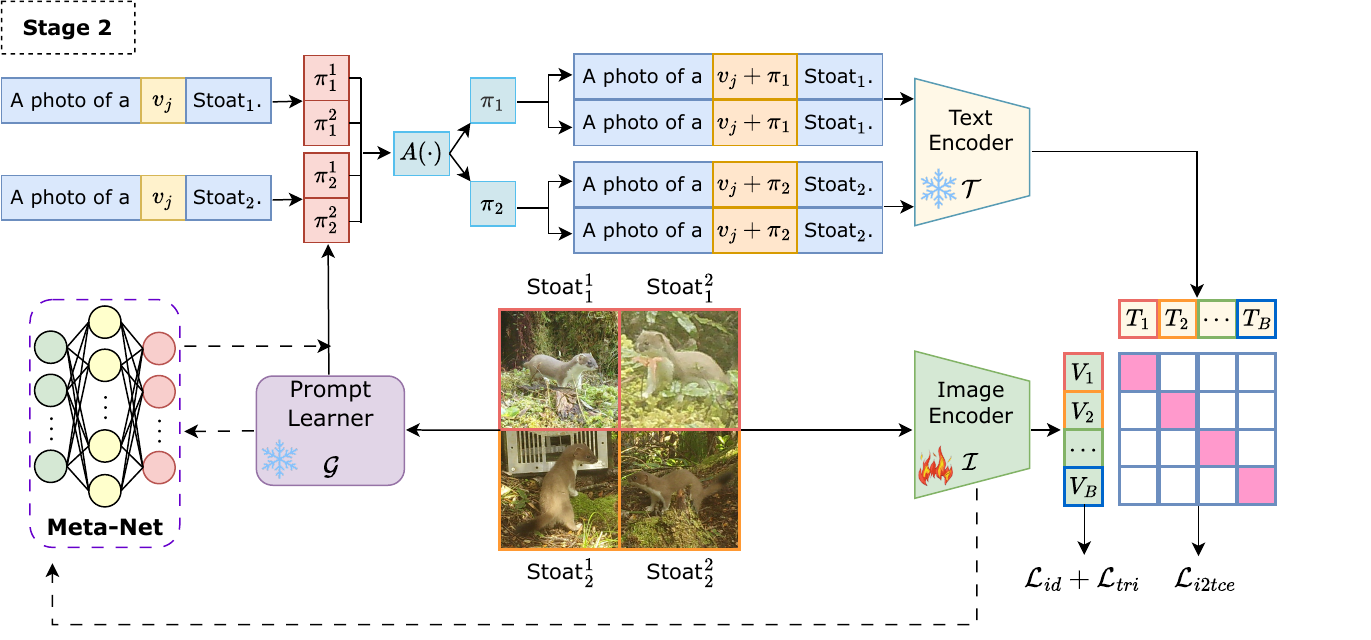}
    \caption{\textbf{Stage 2:} Training of the attention module, $A(\cdot)$, to merge both image-specific and individual-specific descriptions into a uniquely textual description for each individual, accompanied by the image encoder fine-tuning, with a fixed text description generator and text encoder.}
    \label{fig:IndivAID_stage_2_framework}
  \end{subfigure}
  \includegraphics[width=\linewidth]{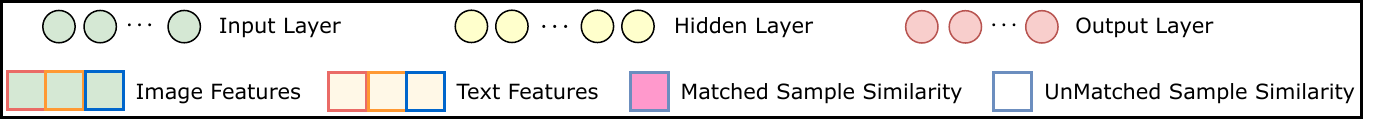}
  \caption{Overview of the \textbf{Indiv}idual \textbf{A}nimal \textbf{ID}entity-Driven (IndivAID) Framework.}
  \label{fig:IndivAID_Framework}
\end{figure*}

\noindent \textbf{Vision-Language Models.} Vision-language models, which align images and texts to create a joint embedding space, encompass two core components: image and text encoders~\cite{radford2021learning, jia2021scaling}.
Historically, image and text encoders are designed and learned independently, later linked by mechanisms for alignment.
Images are often encoded using hand-crafted descriptors~\cite{socher2013zero} or neural networks~\cite{lei2015predicting}, while texts are encoded using pre-trained word vectors~\cite{frome2013devise} or frequency-based TF-IDF features~\cite{elhoseiny2013write}.
Recent advancements in vision-language models involve joint learning of two encoders, often employing larger neural networks.
Notable success in this domain is attributed to advancements in Transformers~\cite{vaswani2017attention}, contrastive representative learning~\cite{henaff2020data}, and access to vast training datasets~\cite{jia2021scaling}.
CLIP~\cite{radford2021learning} exemplifies this, employing two large-scale encoders trained with the contrastive loss on 400 million image-text pairs, showcasing impressive zero-shot recognition capabilities.
CLIP has shown success in various areas, such as semantic segmentation~\cite{rao2022denseclip}, domain adaptation~\cite{bai2024prompt}, domain generalization~\cite{huang2023sentence}, few-shot learning~\cite{zhou2022learning}, and imbalanced learning~\cite{wang2024exploring}.
Despite this, the application of CLIP to Animal ReID remains unexplored.
While efforts have been made to adapt CLIP for Person and Vehicle ReID tasks~\cite{li2023clip}, the challenges in Animal ReID, including pose variations and unconstrained environments, present additional hurdles.

\noindent \textbf{Prompt Learning.} Prompt learning, originating from the NLP domain, leverages pre-trained language models like BERT or GPT as knowledge bases for downstream tasks.
Rather than manual prompt design, prompt learning automates this process using labelled data.
Jiang et al.~\cite{jiang2020can} employed text mining and paraphrasing to generate and select high-performing prompts based on training accuracy.
Shin et al.~\cite{shin2020autoprompt} introduced AutoPrompt, a method that employs a gradient-based strategy to identify tokens that significantly influence model predictions.
Beyond these discrete prompt learning methods, continuous prompt learning has emerged, transforming prompts into learnable vector sets for end-to-end optimization~\cite{lester2021power,li2021prefix}.
In contrast to learning the prompt for CLIP's text encoder, Jia et al.~\cite{jia2022visual} proposed Visual Prompt Learning, which learns a continuous prompt for CLIP's image encoder.
Prompt learning represents an emerging area of study within computer vision, gaining attention only in recent years~\cite{zhou2022conditional,zhou2022learning}.
In this research, we employ continuous prompt learning to bridge the gap between the numeric indices used in Animal ReID tasks and the descriptive labels of image classification.
By generating individual-specific text descriptions for each animal individual, we are able to refine the image encoder, tailoring it specifically for Animal ReID applications.

\section{Method}
\label{sec:method}

\subsection{Preliminaries: Review of CLIP}
\label{sec:preliminaries}

\textbf{Contrastive Language-Image Pre-training (CLIP)}~\cite{radford2021learning} has demonstrated its effectiveness in learning visual concepts from the natural language perspective, revealing its potential in learning open-set visual concepts. CLIP is composed of two fundamental encoders: an image encoder, denoted as $\mathcal{I(\cdot)}$, and a text encoder, denoted as $\mathcal{T(\cdot)}$. Either ResNet~\cite{He_2016_CVPR} or the Vision-Transformer (ViT)~\cite{dosovitskiy2021an} can serve as the image encoder for CLIP. However, our framework relies on ViT for the image encoder since the ViT architecture outperforms ResNet in terms of efficiency and representational capability in CLIP~\cite{radford2021learning}.

The text encoder is also in the structure of a Transformer~\cite{vaswani2017attention}, which converts a sentence of the image description into a textual representation. Typically, for a given sentence of the image description, ``\verb|A photo of a [class].|'', the ``\verb|[class]|'' is replaced by the specific label of the image. Such a description also refers to a Natural Language Processing (NLP) prompt.

CLIP endeavours to learn a multi-modal embedding space by simultaneously training image and text encoders on a batch of $B$ images paired with corresponding texts. Specifically, the batch images can be denoted as $img_i$ with corresponding texts $text_i$, where $i \in \{ 1, 2, \cdots, B \}$. The similarity between $img_i$ and $text_i$ can be computed as
\begin{equation}
    \label{eqn:cos_sim}
    s(V_i, T_i) = V_i \cdot T_i = g_V(img_i) \cdot g_T(text_i),
\end{equation}
where $g_V(\cdot)$ and $g_T(\cdot)$ are linear-projection functions, and $s(\cdot, \cdot)$ is a function to compute the cosine similarity between embeddings in a multi-modal embedding space.

After computing the cosine similarity between embeddings of each image-text pair, the image-to-text contrastive loss $\mathcal{L}_{i2t}$ is calculated as 
\begin{equation}
\label{eqn:i2t_loss}
    \mathcal{L}_{i2t}(i) = -\log \frac{\exp \left( s(V_i, T_i) \right)}{\sum_{a = 1}^{B} \exp \left( s(V_i, T_a) \right)},
\end{equation}
and the text-to-image contrastive loss $\mathcal{L}_{t2i}$ is calculated similarly: 
\begin{equation}
\label{eqn:t2i_loss}
    \mathcal{L}_{t2i}(i) = -\log \frac{\exp \left( s(V_i, T_i) \right)}{\sum_{a = 1}^{B} \exp \left( s(V_a, T_i) \right)},
\end{equation}
where the numerators in Equations~\eqref{eqn:i2t_loss} and~\eqref{eqn:t2i_loss} are the logits (scaled cosine similarity) between embeddings of a matched image-text pair and the denominators are the summation of logits across all texts or images with respect to anchor $V_i$ or $T_i$.

\subsection{IndivAID Framework}
\label{sec:IndivAID}

For most standard classification problems, CLIP converts the concrete labels into textual descriptions, such as ``\verb|A photo of a [class].|'', aiming to align the feature embeddings of images, $V_i$, and texts, $T_i$, derived from the image encoder and text encoder, respectively, in a multi-modal embedding space. However, the labels are indices rather than descriptive words under the ReID setting. It is challenging to directly apply the CLIP framework to Animal ReID tasks due to the lack of semantic labels~\cite{li2023clip}. To overcome the challenges of non-concrete labels and variations in animal images for Animal ReID, we design a two-stage framework, the \textbf{Indiv}idual \textbf{A}nimal \textbf{ID}entity-Driven (IndivAID) framework. As the IndivAID architecture illustrated in Figure~\ref{fig:IndivAID_Framework}, it involves a two-stage training process.

\vspace{5pt}

\noindent \textbf{Stage One Architecture: Design a Text Description Generator.} We propose a \textbf{Text Description Generator}, $\mathcal{G(\cdot)}$, to generate a unique textual description for each image input. Each textual description is composed of static tokens and learnable tokens. Specifically, the textual descriptions that are fed into the text encoder can be formatted as ``\verb|A photo of a| \hspace{0.8pt} $[v_j]$ \hspace{0.8pt} \verb|Animal|$_{id}$.'', where $[v_j]$ is learnable tokens with the same dimension as word embedding, and $j \in (1, 2, \cdots, m)$ denotes $m$ number of learnable textual tokens. We randomly initialize these learnable textual tokens at the beginning of the Stage One training. The term \verb|Animal|$_{id}$ represents various ReID tasks, such as Stoat, Panda, tiger, etc., and the subscript ${id}$ denotes each individual's unique identity. During this stage, we freeze the image and text encoders while training the text description generator to produce an image-specific and individual-specific textual description for each image. Therefore, all the parameters in the image and text encoders remain fixed. Nevertheless, IndivAID attempts to optimize the text description generator to effectively incorporate the meta-tokens relevant to each image. 

A lightweight Meta-Net with the structure (Linear-ReLU-Linear) is adopted to learn the meta-tokens, $\pi_{i}$, where $i \in \{ 1, 2, \cdots, B \}$. Since the meta-tokens correspond to each image, the image embedding extracted from the image encoder is simply the input to the Meta-Net. The dimension of the image embedding is then compressed by $16 \times$ before passing to the hidden layer. The output of the Meta-Net is the learned meta-tokens with the same dimension as word embedding so that IndivAID can comprise the meta-tokens into the learnable textual description associated with each image. As shown in Equation~\eqref{eqn:CARE_stage_1_loss}, we adopt the image-to-text contrastive loss in Equation~\eqref{eqn:i2t_loss}, $\mathcal{L}_{i2t}$, and the text-to-image contrastive loss in Equation~\eqref{eqn:t2i_loss}, $\mathcal{L}_{t2i}$, from the CLIP framework to optimize the text description generator in Stage One. Since we utilize the text description generator to produce the textual descriptions and aim to align the constructed image-specific and individual-specific textual descriptions with the corresponding images, the image-to-text contrastive loss $\mathcal{L}_{i2t}$ and text-to-image contrastive loss $\mathcal{L}_{t2i}$ are used to optimize the text description generator in the Stage One training.
\begin{equation}
    \label{eqn:CARE_stage_1_loss}
    \mathcal{L}_{stage1} = \mathcal{L}_{i2t} + \mathcal{L}_{t2i}
\end{equation}


\noindent \textbf{Stage Two Architecture: Construct Individual-specific Textual Descriptions to Re-identify Animals.} We concentrate on optimizing the attention module, $A(\cdot)$, to combine both image-specific and individual-specific textual descriptions into a uniquely textual description for each individual, accompanied by the adaption of the image encoder while freezing the text description generator and text encoder. To align with the standard object ReID frameworks, we employ the identity loss~\cite{luo2019bag}, $\mathcal{L}_{id}$, with a Label Smoothing Regularization (LSR)~\cite{Szegedy2016rethinking} attached and the triplet loss~\cite{luo2019bag, hermans2017defense}, $\mathcal{L}_{tri}$. The LSR is applied to the identity loss with a smoothing parameter, $\epsilon$, to prevent overfitting and make the model more generalizable. The identity loss with a LSR can be formulated as 
\begin{equation}
\label{eqn:id_loss}
    \mathcal{L}_{id} = \sum_{k = 1}^{N} -q_k \log(p_k), 
\end{equation}
where $q_k = (1 - \epsilon) \delta_{k, y} + \epsilon / N$ denotes the value in the target distribution, and $p_k$ is the identity prediction probability of a class $k$. Equation~\eqref{eqn:delta} shows the details of $\delta_{k, y}$. Meanwhile, the formulae of $p_k$ is demonstrated as Equation~\eqref{eqn:id_pred_prob}. The value $q_k$ is derived from the target distribution, $q\left(k|x\right)$, as shown in Equation~\eqref{eqn:target_dist}, with weights $1 - \epsilon$ and $\epsilon$ assigned to the original ground-truth distribution and fixed distribution, respectively. In our experiment setting, we adopt a uniform distribution, $U(k) = 1 / N$, to align with the works of Szegedy et al.~\cite{Szegedy2016rethinking} and Li et al.~\cite{li2023clip}.

\begin{equation}
\label{eqn:delta}
    \delta_{k, y} = 
    \begin{cases}
        1, \; \text{if} \; k = y;\\
        0, \; \text{otherwise}.
    \end{cases}
\end{equation}

\begin{equation}
\label{eqn:id_pred_prob}
    p_k = \mathbb{P}\left(k | x\right) = \frac{\exp (z_k)}{\sum_{i = 1}^{N} \exp (z_i)}, 
\end{equation}
where $z_i$ denotes the logits of a class $i$ and $k \in \{ 1, 2, \cdots, N \}$.

\begin{equation}
\label{eqn:target_dist}
    q\left(k|x\right) = (1 - \epsilon) \cdot \delta_{k, y|x} + \epsilon \cdot U(k), 
\end{equation}
where $U(k)$ denotes a fixed distribution over labels, which is independent of the training data, $x$.

\noindent The triplet loss is formulated as 
\begin{equation}
    \label{eqn:triplet_loss}
    \mathcal{L}_{tri} = \left[ d_p - d_n + \tau \right]_{+}, 
\end{equation}
where $d_p$ and $d_n$ denote the feature distances of the positive pair and the negative pair, respectively, $\tau$ is a margin hyperparameter ensuring that the feature distance between an anchor $x_i$ and a positive example $x_p$ is smaller than that between the anchor $x_i$ and a negative example $x_n$ by at least $\tau$, preventing the case when $d_p = d_n$, and $[\cdot]_{+} = \max(0, \cdot)$. By relying on the inspiration of nearest-neighbour, the triplet loss is designed to make all points with the same identity closer than those with different identities, thus forming a cluster of points for each identity. Nevertheless, the formulation of the triplet loss is only based on the distance differences rather than on the magnitude of the differences. For instance, if $d_p = 0.5$ and $d_n = 0.8$ given the margin $\tau = 0.4$, the triplet loss is $0.1$. Similarly, if $d_p = 1.5$ and $d_n = 1.8$ given the margin $\tau = 0.4$, the triplet loss is $0.1$.

To adopt and maximize CLIP's potential on Animal ReID tasks, we also utilize an image-to-text cross-entropy loss, $\mathcal{L}_{i2tce}$, as illustrated in Equation~\eqref{eqn:i2t_ce_loss}. For each image, $x_i$, we need to compute $\mathcal{L}_{i2tce}$ by comparing the image feature embedding with all the text feature embeddings from those individual-specific textual descriptions within a cross-modal embedding space. Analogous to $\mathcal{L}_{id}$, LSR is applied to $q_k$ in $\mathcal{L}_{i2tce}$.

\begin{equation}
\label{eqn:i2t_ce_loss}
    \mathcal{L}_{i2tce}(i) = \sum_{k = 1}^{N} - q_k \log \frac{\exp \left( s(V_i, T_{y_k}) \right)}{\sum_{y_a = 1}^{N} \exp \left( s(V_i, T_{y_a}) \right)}.
\end{equation}

Ultimately, as demonstrated in Equation~\eqref{eqn:CARE_stage_2_loss}, Stage 2 loss can be formulated as 
\begin{equation}
    \label{eqn:CARE_stage_2_loss}
    \mathcal{L}_{stage2} = \mathcal{L}_{id} + \mathcal{L}_{tri} + \mathcal{L}_{i2tce}.
\end{equation}

\begin{figure}[!ht]
    \centering
    \includegraphics[width=\columnwidth]{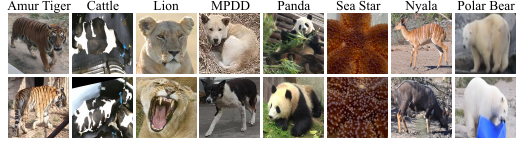}
    \caption{Examples of Benchmark Datasets.}
    \label{fig:benchmark}
\end{figure}

\section{Experiments}
\label{sec:experiments}

\begin{table*}[!ht]
\centering
\resizebox{\linewidth}{!}{
\begin{tabular}{@{}l|rrrrrrrr@{}}
\toprule
\multicolumn{1}{c|}{\textbf{Dataset}} & \textbf{Train (Imgs)} & \textbf{Train (IDs)} & \textbf{Gallery (Imgs)} & \textbf{Gallery (IDs)} & \textbf{Query (Imgs)} & \textbf{Query (IDs)} & \textbf{Total (Imgs)} & \textbf{Total (IDs)} \\ \midrule
ATRW               & 3,730   & 149 & 521   & 33 & 424    & 33 & 4,675   & 182 \\ 
FriesianCattle2017 & 752     & 66  & 97    & 18 & 85     & 18 & 934     & 84  \\ 
LionData           & 594     & 75  & 85    & 19 & 61     & 19 & 740     & 94  \\ 
MPDD               & 1,032   & 95  & 521   & 96 & 104    & 96 & 1,657   & 191 \\ 
IPanda50           & 5,620   & 40  & 677   & 10 & 577    & 10 & 6,874   & 50  \\ 
SeaStarReID2023    & 1,768   & 81  & 230   & 14 & 189    & 14 & 2,187   & 95  \\ 
NyalaData          & 1,213   & 179 & 375   & 58 & 354    & 58 & 1,942   & 237 \\ 
PolarBearVidID     & 107,250 & 10  & 26,500 & 3  & 4,613 & 3  & 138,363 & 13  \\ 
\bottomrule
\end{tabular}
}
\caption{Details of Benchmark Datasets.}
\label{tab:benchmark_detail}
\end{table*}

\subsection{Experimental Setting}
\label{sec:experimental_setting}

\noindent \textbf{Datasets.} We evaluated the IndivAID framework on eight Animal ReID benchmarks: \textbf{ATRW}~\cite{li2020atrw}, \textbf{FriesianCattle2017}~\cite{andrew2017visual}, \textbf{LionData}~\cite{dlamini2020automated}, \textbf{MPDD}~\cite{he2023animal}, \textbf{IPanda50}~\cite{wang2021giant}, \textbf{SeaStarReID2023}~\cite{wahltinez2024open}, \textbf{NyalaData}~\cite{dlamini2020automated}, and \textbf{PolarBearVidID}~\cite{zuerl2023polarbearvidid}, comprising the data with different characteristics and complexity levels. The statistics of each benchmark dataset are summarized in Table~\ref{tab:benchmark_detail} and Figure~\ref{fig:benchmark} shows examples of each benchmark dataset.

\vspace{5pt}

\noindent \textbf{Evaluation Metrics.} We utilized two commonly used metrics in ReID tasks to evaluate the performance: mean Average Precision (mAP)~\cite{wu2017rgbinfrared} and Cumulative Matching Characteristics (CMC)~\cite{wu2017rgbinfrared, hyeonjoon2001computational}. mAP evaluates the overall retrieval performance by calculating the average precision across all queries. CMC (Top-$k$ accuracy) measures the model's ability to correctly identify the match within the top-$k$ retrieved images from the Gallery Set. Specifically, the mAP metric concentrates on evaluating the model's overall performance across all query images, but the Top-$k$ accuracy reveals the number of images that have to be examined to achieve a desired level of performance. Both mAP and Top-$k$ accuracy computations rely on the cosine similarity metric.

\vspace{5pt}

\begin{table*}[!ht]
\centering
\resizebox{\linewidth}{!}{%
\begin{tabular}{@{}lrrrr|lrrrr@{}}
\toprule
 & \multicolumn{4}{c}{\textbf{ATRW}} & & \multicolumn{4}{c}{\textbf{FriesianCattle2017}} \\ \midrule
 & mAP & Top-1 & Top-5 & Top-10 & & mAP & Top-1 & Top-5 & Top-10 \\ \midrule
 CLIP-ZS & 40.25 ± .00 & 86.32 ± .00 & 96.23 ± .00 & 98.35 ± .00 &
 CLIP-ZS & 57.76 ± .00 & 76.47 ± .00 & 89.41 ± .00 & 94.12 ± .00 \\
 CLIP-FT & 53.10 ± .17 & 91.05 ± .48 & 96.73 ± .20 & 98.31 ± .34 &
 CLIP-FT & 83.65 ± .28 & 93.65 ± .42 & 100.00 ± .00 & 100.00 ± .00 \\
 CLIP-ReID & 53.72 ± .24 & 93.91 ± .31 & 98.14  ± .46 & 98.37  ± .49 &
 CLIP-ReID & 87.32 ± .69 & 97.56 ± .78 & 100.00 ± .00 & 100.00 ± .00 \\
 IndivAID      & \textbf{55.77 ± .21} & \textbf{95.81 ± .14} & 98.84  ± .31 & \textbf{99.52  ± .27} &
 IndivAID      & \textbf{93.66 ± .10} & \textbf{99.23 ± .16} & 100.00 ± .00 & 100.00 ± .00\\ \midrule
 & \multicolumn{4}{c}{\textbf{LionData}} & & \multicolumn{4}{c}{\textbf{MPDD}} \\ \midrule
 & mAP & Top-1 & Top-5 & Top-10 & & mAP & Top-1 & Top-5 & Top-10 \\ \midrule
 CLIP-ZS & 13.60 ± .00 & 16.39 ± .00 & 40.98 ± .00 & 57.38 ± .00 &
 CLIP-ZS & 53.21 ± .00 & 68.27 ± .00 & 87.50 ± .00 & 95.19 ± .00 \\
 CLIP-FT & 21.63 ± .33 & 26.23 ± .28 & 62.30 ± .15 & 75.96 ± .40 &
 CLIP-FT & 81.10 ± .31 & 90.12 ± .53 & 99.04 ± .09 & 99.14 ± .07 \\
 CLIP-ReID & 25.61 ± .47 & 31.73 ± .43 & 71.61 ± .30 & 82.54 ± .19 &
 CLIP-ReID & 84.91 ± .22 & 96.23 ± .18 & 99.04 ± .17 & 99.31 ± .40 \\ 
 IndivAID      & \textbf{39.56 ± .09} & \textbf{56.77 ± .25} & \textbf{85.04 ± .29} & \textbf{93.24 ± .38} &
 IndivAID      & \textbf{88.24 ± .15} & \textbf{98.14 ± .25} & 99.32 ± .50 & 99.74 ± .36 \\ \midrule
 & \multicolumn{4}{c}{\textbf{IPanda50}} & & \multicolumn{4}{c}{\textbf{SeaStarReID2023}} \\ \midrule
 & mAP & Top-1 & Top-5 & Top-10 & & mAP & Top-1 & Top-5 & Top-10 \\ \midrule
 CLIP-ZS & 17.01 ± .00 & 29.64 ± .00 & 64.99 ± .00 & 78.86 ± .00 &
 CLIP-ZS & 20.60 ± .00 & 45.50 ± .00 & 79.89 ± .00 & 91.01 ± .00 \\
 CLIP-FT & 26.86 ± .12 & 41.36 ± .27 & 75.16 ± .55 & 87.35 ± .31 &
 CLIP-FT & 65.38 ± .49 & 89.07 ± .37 & 98.41 ± .24 & 100.00 ± .00 \\
 CLIP-ReID & 29.73 ± .21 & 42.14 ± .11 & 75.87  ± .14 & 88.76  ± .56 &
 CLIP-ReID & 78.73 ± .38 & 96.82 ± .36 & 100.00 ± .00 & 100.00 ± .00\\ 
 IndivAID      & \textbf{36.56 ± .40} & \textbf{48.63 ± .15} & \textbf{82.02 ± .16} & \textbf{91.47 ± .26} &
 IndivAID      & \textbf{83.76 ± .17} & \textbf{97.94 ± .34} & 100.00 ± .00 & 100.00 ± .00 \\ \midrule
 & \multicolumn{4}{c}{\textbf{NyalaData}} & & \multicolumn{4}{c}{\textbf{PolarBearVidID}} \\ \midrule
 & mAP & Top-1 & Top-5 & Top-10 & & mAP & Top-1 & Top-5 & Top-10 \\ \midrule
 CLIP-ZS &  9.55 ± .00 & 15.54 ± .00 & 38.70 ± .00 & 53.95 ± .00 &
 CLIP-ZS & 39.46 ± .00 & 80.23 ± .00 & 88.45 ± .00 & 91.00 ± .00 \\
 CLIP-FT & 10.89 ± .15 & 18.93 ± .37 & 45.29 ± .22 & 61.96 ± .16 &
 CLIP-FT & 43.19 ± .18 & 80.44 ± .48 & 89.20 ± .33 & 91.18 ± .40 \\
 CLIP-ReID & 13.44 ± .13 & 21.92 ± .27 & 51.00 ± .46 & 67.45 ± .29 &
 CLIP-ReID & 44.84 ± .47 & 81.29 ± .22 & 90.23 ± .16 & 91.92 ± .19 \\
 IndivAID      & \textbf{16.25 ± .09} & \textbf{27.63 ± .19} & \textbf{61.04 ± .20} & \textbf{76.31 ± .18} &
 IndivAID      & \textbf{47.73 ± .26} & \textbf{83.64 ± .17} & \textbf{92.32 ± .21} & \textbf{94.69 ± .23} \\ 
\bottomrule
\end{tabular}
}
\caption{Evaluation Results on Eight Benchmark Datasets.}
\label{tab:evaluation_results}
\end{table*}

\noindent \textbf{Comparisons with Baseline Methods.} We evaluated our IndivAID framework by comparing the performance of some baseline methods on eight benchmark datasets. By adopting the vision-language model CLIP, \textbf{CLIP-ReID}~\cite{li2023clip} is the state-of-the-art method designed for Person and Vehicle ReID tasks, which outperforms all the old CNN- and ViT-based methods by a significant margin. CLIP-ReID proposed uniquely producing a textual description for each individual without incorporating the image features into the construction of the textual description. Nonetheless, IndivAID proposes a text description generator to produce an image-specific and individual-specific textual description for each image, which integrates the image features into generating the textual descriptions. Furthermore, IndivAID utilizes an attention module to merge each individual's image-specific textual descriptions into a uniquely individual-specific textual description. Due to the challenges of the pose variations and unconstrained environments in Animal ReID tasks, IndivAID concentrates on attaching the image features to the textual descriptions by exploiting the potential of the vision-language model to improve the performance of Animal ReID. Therefore, we can demonstrate the effectiveness of the proposed text description generator on Animal ReID tasks through comparisons with the CLIP-ReID method.

Additionally, we assessed a baseline variant, \textbf{CLIP Fine-Tuning (CLIP-FT)}, which involves fine-tuning the CLIP's image encoder straightforwardly without incorporating the learnable textual descriptions produced by the text description generator. Specifically, CLIP-FT utilizes initialized concrete textual descriptions and does not engage in Stage One learning but only fine-tunes the CLIP's image encoder as in Stage Two. Meanwhile, a stoat image is passed to CLIP's image encoder, and CLIP-FT fine-tunes the image encoder by adopting IndivAID's Stage 2 loss, as demonstrated in Equation~\eqref{eqn:CARE_stage_2_loss}. Thus, we can illustrate the superiority of the learnable textual descriptions on Animal ReID tasks by comparing the CLIP-FT baseline.

Finally, we examined the zero-shot ReID capabilities of CLIP, \textbf{CLIP-ZeroShot (CLIP-ZS)}, which does not involve any training but utilizes all the frozen parameters in CLIP's image encoder to make the inferences. However, the textual descriptions and CLIP's text encoder completely do not participate in the CLIP-ZS baseline. Hence, CLIP-ZS can prove the potential of the CLIP on Animal ReID tasks even without training or fine-tuning.

\vspace{5pt}

\noindent \textbf{Network Structure.} IndivAID utilizes CLIP's image $\mathcal{I}(\cdot)$ and text encoders $\mathcal{T}(\cdot)$ as the fundamental components for image and text feature extraction. Given CLIP's options for its image encoder, a Vision-Transformer (ViT), or a CNN with global attention pooling, we prefer the ViT architecture due to its proficiency in capturing long-range dependencies in images. Specifically, we employ the ViT-B/16 variant, featuring 12 Transformer layers with a 768-dimensional hidden size. To align with the text encoder's output, we employ a linear layer to reduce the image feature vector dimensionality from 768 to 512.

\vspace{5pt}

\noindent \textbf{Reproducibility.} Our methodology is implemented using the PyTorch libraries. All experiments are conducted on NVIDIA Tesla A100 GPUs. In the first stage, we utilize the Adaptive Moment Estimation (Adam)~\cite{kingma2014adam} optimizer with a learning rate initialized at 3.5 $\times$ $10^{-4}$ and decayed by a cosine schedule. The batch size is set to 1 for all benchmarks without any augmentation methods. The number of learnable tokens in textual descriptions is set to be 4, and these learnable textual descriptions are initialized with the pre-trained word embeddings of ``\verb|A photo of a.|''.

In the second stage, we also use the Adam optimizer to fine-tune the image encoder and train the attention module. The batch size $B$ depends on the number of individuals, $I$, within a batch and the number of images, $K$, selected per individual, where $B = I \times K$. Each image is augmented by random horizontal flipping, padding, cropping, and erasing. We warm up the model for ten epochs with a linearly growing learning rate from $5\times 10^{-7}$ to $5\times 10^{-6}$. Then, the learning rate is decayed by a factor of 0.1. The margin hyperparameter $\tau$ for the triplet loss is set to 0.3.

\subsection{Experimental Results}
\label{sec:experimental_results}

Table~\ref{tab:evaluation_results} shows the evaluation results of IndivAID and all baseline methods, CLIP-ZS, CLIP-FT, and CLIP-ReID, on eight benchmark datasets. All mAP, Top-1, Top-5, and Top-10 Accuracy are averaged across ten runs with 95\% confidence intervals. As discussed in Section~\ref{sec:experimental_setting}, CLIP-ZS stands for zero-shot learning with the vision-language model CLIP. However, the CLIP-ZS results on NyalaData and PolarBearVidID reveal the effectiveness of CLIP's image encoder, as the difference in mean Average Precision (mAP) between CLIP-ZS and IndivAID, which achieves the highest mAP on both datasets, is approximately 7\% and 8\% on NyalaData and PolarBearVidID, respectively. Therefore, these results verify the potential of CLIP on Animal ReID tasks even without involving any training.

CLIP-FT stands for the CLIP Fine-Tuning baseline, which involves fine-tuning the CLIP's image encoder and incorporating concrete textual descriptions. The results of CLIP-FT are significantly higher than CLIP-ZS across all evaluation metrics. By comparing the results between CLIP-FT and CLIP-ZS, we noticed that the mAP and Top-1 Accuracy of CLIP-FT on SeaStarReID2023 exceedingly surpass the CLIP-ZS by about 45\% and 44\%, respectively. Meanwhile, CLIP-FT achieves approximately 26\% higher mAP in the FriesianCattle2017 dataset and around 28\% higher mAP in the MPDD dataset. These superior results further demonstrate the effectiveness of textual descriptions on Animal ReID tasks with the adaptation of CLIP.

The state-of-the-art method, CLIP-ReID, designed for Person and Vehicle ReID tasks, extends the CLIP-FT baseline by incorporating the learnable textual tokens to produce individual-specific textual descriptions. CLIP-ReID outperforms CLIP-FT across all eight benchmark datasets on mAP and Top-1 Accuracy metrics. Specifically, the mAP of CLIP-ReID on SeaStarReID2023 exceeds the CLIP-FT by approximately 13\%. Thus, CLIP-ReID illustrates the efficacy of uniquely constructing a textual description for each individual in Animal ReID tasks, bridging the gap of non-descriptive labels in ReID tasks.

In addition, the evaluation results indicate the superior performance of IndivAID compared to CLIP-ZS, CLIP-FT, and the state-of-the-art method, CLIP-ReID, across all eight benchmark datasets. IndivAID consistently surpasses all compared methods in each metric and dataset. Both CLIP-ZS and CLIP-FT exhibit local optimal results, underscoring the limitations of CLIP in ReID tasks without descriptive text labels. Conversely, CLIP-ReID and IndivAID demonstrate superior performance by utilizing CLIP's text encoder capabilities through learned textual descriptions. LionData and IPanda50 are challenging datasets due to the relatively low mAP. Nevertheless, IndivAID achieves approximately 10\% higher mAP in the LionData and IPanda50 datasets and about 20\% higher Top-1 Accuracy in the LionData dataset than the CLIP-ReID method. Hence, the evaluation results on eight benchmark datasets reveal the effectiveness of IndivAID's text description generator, which produces individual-specific textual descriptions incorporating the image feature from each image input in Animal ReID tasks.

\subsection{Ablation Study}
\label{sec:ablation_study}

\begin{table}[!ht]
\centering
\begin{tabular}{@{}c|c|c|r|r@{}}
\toprule
 \multicolumn{5}{c}{\textbf{NyalaData}} \\ \midrule
 $\mathcal{L}_{i2tce}$ & $\mathcal{L}_{i2t}$ & $\mathcal{L}_{t2i}$ & mAP & Top-1 \\ \midrule
 - & $\checkmark$ & $\checkmark$ & 15.62 ± .14 & 26.48 ± .13 \\
 - & $\checkmark$ & - & 16.00 ± .18 & 26.85 ± .08 \\
 $\checkmark$ & - & $\checkmark$ & 16.07 ± .06 & 27.21 ± .14 \\
 $\checkmark$ & - & - & \textbf{16.25 ± .09} & \textbf{27.63 ± .19} \\
\bottomrule
\end{tabular}
\caption{The mAP and Top-1 Accuracy of IndivAID on the NyalaData benchmark dataset with various contrastive loss combinations in Stage Two training.}
\label{tab:contrastive_loss_analysis}
\end{table}

\noindent \textbf{Ablation Study on the Contrastive loss functions in Stage Two Training.} As illustrated in Equation~\eqref{eqn:CARE_stage_2_loss}, we employ the identity loss, $\mathcal{L}_{id}$, triplet loss, $\mathcal{L}_{tri}$, and image-to-text cross-entropy loss, $\mathcal{L}_{i2tce}$, to optimize the attention module and CLIP's image encoder in Stage Two training. The identity loss and triplet loss are standard loss functions for ReID tasks. Moreover, IndivAID incorporates the image-to-text cross-entropy loss to maximize the potential of CLIP in Animal ReID. To investigate the sensitivity of our proposed IndivAID framework, we studied the effects of utilizing different contrastive loss functions in the Stage Two training phase on the Animal ReID performance.

As shown in Table~\ref{tab:contrastive_loss_analysis}, we conducted an ablation study on the NyalaData benchmark dataset to investigate the effects of various combinations of contrastive losses, such as image-to-text cross-entropy loss, $\mathcal{L}_{i2tce}$, image-to-text contrastive loss, $\mathcal{L}_{i2t}$, and text-to-image contrastive loss, $\mathcal{L}_{t2i}$. Table~\ref{tab:contrastive_loss_analysis} reveals that incorporating the text-to-image contrastive loss, $\mathcal{L}_{t2i}$, into the Stage Two training degrades the model's performance. For example, the combination of $\mathcal{L}_{i2t}$ and $\mathcal{L}_{t2i}$ has lower mAP and Top-1 Accuracy than the $\mathcal{L}_{i2t}$. Similarly, the mAP and Top-1 Accuracy of IndivAID with an image-to-text cross-entropy loss, $\mathcal{L}_{i2tce}$, surpass the combination of $\mathcal{L}_{i2tce}$ and $\mathcal{L}_{t2i}$. Therefore, adopting the text-to-image contrastive loss, $\mathcal{L}_{t2i}$, to the Stage Two training degrades the model's performance on Animal ReID. As discussed in Section~\ref{sec:IndivAID}, IndivAID aims to optimize the attention module and fine-tune the CLIP's image encoder but fix the text description generator and CLIP's text encoder in Stage Two, so the text-to-image contrastive loss is unnecessary for the Stage Two training.

Furthermore, we realized that utilizing the image-to-text cross-entropy loss, $\mathcal{L}_{i2tce}$, in Stage Two training is more effective than the image-to-text contrastive loss, $\mathcal{L}_{i2t}$. As illustrated in Equations~\eqref{eqn:i2t_loss} and \eqref{eqn:i2t_ce_loss}, the image-to-text contrastive loss, $\mathcal{L}_{i2t}$, only considers text features for the identities within a batch $B$. However, the image-to-text cross-entropy loss, $\mathcal{L}_{i2tce}$, considers text features for all the identities in the training dataset. Table~\ref{tab:contrastive_loss_analysis} indicates that the IndivAID framework with the image-to-text cross-entropy loss in Stage Two is superior to the image-to-text contrastive loss.

\section{Case Study: Stoat ReID}
\label{sec:Stoat_ReID}

New Zealand faces a significant ecological challenge, with an estimated 68,000 native birds falling prey to invasive species like stoats, possums, rats, and mustelids nightly~\cite{DOC}. The removal of these predators is crucial for the survival and prosperity of endemic species. Therefore, the New Zealand government started the Predator Free 2050 project, which focuses on completely removing five predators: rats, stoats, ferrets, weasels, and possums. Our IndivAID framework aims to support this goal by helping stoat eradication on Waiheke Island.

\subsection{Datasets}

Data collection and annotation from the wild is challenging. Our dataset comprises images from two locations, Waiheke Island and the South Island of New Zealand, enriching the diversity of data. The Waiheke Island dataset contains 13 images with 5 individuals. The South Island dataset contains 183 images with 56 individuals. Due to the extensive small dataset, we aim to use the South Island dataset to train the model and evaluate using the Waiheke Island dataset. However, training the model with South Island data and evaluating it on Waiheke Island data introduces a distribution shift, typically resulting in performance degradation~\cite{li2023clip, zhao2024symmetric}.

\begin{table}[!ht]
\centering
\resizebox{\columnwidth}{!}{%
\begin{tabular}{@{}lcccc@{}}
\toprule
& \multicolumn{4}{c}{Stoat} \\ \midrule
 & mAP & Top-1 & Top-5 & Top-10 \\ \midrule
CLIP-ZS & 43.73 ± .00 & 61.54 ± .00 & 84.62 ± .00 & 100.00 ± .00 \\
CLIP-FT & 45.45 ± .39 & 56.41 ± .21 & 89.13 ± .44 & 100.00 ± .00\\
CLIP-ReID & 45.84 ± .27 & 63.89 ± .77 & 92.31 ± .34 & 100.00 ± .00 \\
IndivAID      & \textbf{52.54 ± .30} & \textbf{71.82 ± .41} & \textbf{97.26 ± .69} & 100.00 ± .00 \\ 
\bottomrule
\end{tabular}
}
\caption{Evaluation Results on Stoat Dataset.}
\label{tab:stoat_results}
\end{table}

\subsection{Experimental Results}

Table~\ref{tab:stoat_results} displays the evaluation results of IndivAID compared to baselines on the Stoat dataset. Similar to the prior discussions in Section~\ref{sec:experimental_results}, IndivAID consistently outperforms all baselines, particularly in mAP and Top-1 Accuracy. These results underscore two key insights. Firstly, IndivAID exhibits superior performance on the real-world dataset, illustrating its practical applicability. Secondly, despite the distribution shift between the training (South Island) and test (Waiheke Island) datasets, IndivAID's performance highlights its robustness to distribution shifts - a prevalent challenge in real-world ReID problems.

\section{Conclusion}
\label{sec:conclusion}

By leveraging the vision-language model CLIP, we proposed a two-stage framework, the \textbf{Indiv}idual \textbf{A}nimal \textbf{ID}entity-Driven (IndivAID) framework, to overcome the challenge of non-concrete labels in Animal ReID tasks. IndivAID concentrates on attaching the image features to the textual descriptions by exploiting the potential of the vision-language model to address the challenge of variations in animal images, improving the performance of Animal ReID. Specifically, in Stage One, IndivAID trains a text description generator to produce image-specific and individual-specific textual descriptions by considering each image within individuals. During the Stage Two training, IndivAID fine-tunes CLIP's image encoder by incorporating the individual-specific textual descriptions merged with an attention module. The superior evaluation results on eight benchmark datasets and a real-world Stoat dataset reveal the effectiveness and applicability of IndivAID in Animal ReID.

{
    \small
    \bibliographystyle{ieeenat_fullname}
    \bibliography{main}
}


\end{document}